\documentclass[10pt,twocolumn,letterpaper]{article}
\usepackage{wacv}
\usepackage{times}
\usepackage{epsfig}
\usepackage{graphicx}
\usepackage{amsmath}
\usepackage{amssymb}
\usepackage{times}
\usepackage{float}
\usepackage{enumerate,booktabs}
\usepackage[numbers,sort]{natbib}
\usepackage{caption}
\usepackage{subcaption}
\usepackage{multirow,bigdelim,dcolumn,booktabs}
\usepackage{hhline}
\usepackage[export]{adjustbox}

\wacvfinalcopy 

\def\wacvPaperID{241} 

\ifwacvfinal\fi
\begin{document}
\def\mA{\mathcal{A}}
\def\mB{\mathcal{B}}
\def\mC{\mathcal{C}}
\def\mD{\mathcal{D}}
\def\mE{\mathcal{E}}
\def\mF{\mathcal{F}}
\def\mG{\mathcal{G}}
\def\mH{\mathcal{H}}
\def\mI{\mathcal{I}}
\def\mJ{\mathcal{J}}
\def\mK{\mathcal{K}}
\def\mL{\mathcal{L}}
\def\mM{\mathcal{M}}
\def\mN{\mathcal{N}}
\def\mO{\mathcal{O}}
\def\mP{\mathcal{P}}
\def\mQ{\mathcal{Q}}
\def\mR{\mathcal{R}}
\def\mS{\mathcal{S}}
\def\mT{\mathcal{T}}
\def\mU{\mathcal{U}}
\def\mV{\mathcal{V}}
\def\mW{\mathcal{W}}
\def\mX{\mathcal{X}}
\def\mY{\mathcal{Y}}
\def\mZ{\mathcal{Z}}

\def\1n{\mathbf{1}_n}
\def\0{\mathbf{0}}
\def\1{\mathbf{1}}

\def\A{{\bf A}}
\def\B{{\bf B}}
\def\C{{\bf C}}
\def\D{{\bf D}}
\def\E{{\bf E}}
\def\F{{\bf F}}
\def\G{{\bf G}}
\def\H{{\bf H}}
\def\I{{\bf I}}
\def\J{{\bf J}}
\def\K{{\bf K}}
\def\L{{\bf L}}
\def\M{{\bf M}}
\def\N{{\bf N}}
\def\O{{\bf O}}
\def\P{{\bf P}}
\def\Q{{\bf Q}}
\def\R{{\bf R}}
\def\S{{\bf S}}
\def\T{{\bf T}}
\def\U{{\bf U}}
\def\V{{\bf V}}
\def\W{{\bf W}}
\def\X{{\bf X}}
\def\Y{{\bf Y}}
\def\Z{{\bf Z}}

\def\a{{\bf a}}
\def\b{{\bf b}}
\def\c{{\bf c}}
\def\d{{\bf d}}
\def\e{{\bf e}}
\def\f{{\bf f}}
\def\g{{\bf g}}
\def\h{{\bf h}}
\def\i{{\bf i}}
\def\j{{\bf j}}
\def\k{{\bf k}}
\def\l{{\bf l}}
\def\m{{\bf m}}
\def\n{{\bf n}}
\def\o{{\bf o}}
\def\p{{\bf p}}
\def\q{{\bf q}}
\def\r{{\bf r}}
\def\s{{\bf s}}
\def\t{{\bf t}}
\def\u{{\bf u}}
\def\v{{\bf v}}
\def\w{{\bf w}}
\def\x{{\bf x}}
\def\y{{\bf y}}
\def\z{{\bf z}}

\def\balpha{\mbox{\boldmath{$\alpha$}}}
\def\bbeta{\mbox{\boldmath{$\beta$}}}
\def\bdelta{\mbox{\boldmath{$\delta$}}}
\def\bgamma{\mbox{\boldmath{$\gamma$}}}
\def\blambda{\mbox{\boldmath{$\lambda$}}}
\def\bsigma{\mbox{\boldmath{$\sigma$}}}
\def\btheta{\mbox{\boldmath{$\theta$}}}
\def\bomega{\mbox{\boldmath{$\omega$}}}
\def\bxi{\mbox{\boldmath{$\xi$}}}
\def\bnu{\mbox{\boldmath{$\nu$}}}                                  
\def\bphi{\mbox{\boldmath{$\phi$}}}

\def\bDelta{\mbox{\boldmath{$\Delta$}}}
\def\bOmega{\mbox{\boldmath{$\Omega$}}}
\def\bPhi{\mbox{\boldmath{$\Phi$}}}
\def\bLambda{\mbox{\boldmath{$\Lambda$}}}
\def\bSigma{\mbox{\boldmath{$\Sigma$}}}
\def\bGamma{\mbox{\boldmath{$\Gamma$}}}

\newcommand{\myminimum}[1]{\mathop{\textrm{minimum}}_{#1}}
\newcommand{\mymaximum}[1]{\mathop{\textrm{maximum}}_{#1}}    
\newcommand{\mymin}[1]{\mathop{\textrm{minimize}}_{#1}}
\newcommand{\mymax}[1]{\mathop{\textrm{maximize}}_{#1}}
\newcommand{\mymins}[1]{\mathop{\textrm{min.}}_{#1}}
\newcommand{\mymaxs}[1]{\mathop{\textrm{max.}}_{#1}}  
\newcommand{\myargmin}[1]{\mathop{\textrm{argmin}}_{#1}} 
\newcommand{\myargmax}[1]{\mathop{\textrm{argmax}}_{#1}} 
\newcommand{\myst}{\textrm{s.t. }}

\newcommand{\denselist}{\itemsep -1pt}
\newcommand{\sparselist}{\itemsep 1pt}

\definecolor{pink}{rgb}{0.9,0.5,0.5}
\definecolor{purple}{rgb}{0.5, 0.4, 0.8}   
\definecolor{gray}{rgb}{0.3, 0.3, 0.3}
\definecolor{mygreen}{rgb}{0.2, 0.6, 0.2}

\newcommand{\cyan}[1]{\textcolor{cyan}{#1}}
\newcommand{\red}[1]{\textcolor{red}{#1}}  
\newcommand{\blue}[1]{\textcolor{blue}{#1}}
\newcommand{\magenta}[1]{\textcolor{magenta}{#1}}
\newcommand{\pink}[1]{\textcolor{pink}{#1}}
\newcommand{\green}[1]{\textcolor{green}{#1}} 
\newcommand{\gray}[1]{\textcolor{gray}{#1}}    
\newcommand{\mygreen}[1]{\textcolor{mygreen}{#1}}    
\newcommand{\purple}[1]{\textcolor{purple}{#1}}       

\definecolor{greena}{rgb}{0.4, 0.5, 0.1}
\newcommand{\greena}[1]{\textcolor{greena}{#1}}

\definecolor{bluea}{rgb}{0, 0.4, 0.6}
\newcommand{\bluea}[1]{\textcolor{bluea}{#1}}
\definecolor{reda}{rgb}{0.6, 0.2, 0.1}
\newcommand{\reda}[1]{\textcolor{reda}{#1}}

\def\changemargin#1#2{\list{}{\rightmargin#2\leftmargin#1}\item[]}
\let\endchangemargin=\endlist
                                               
\newcommand{\cm}[1]{}

\title{X-ray Scattering Image Classification Using Deep Learning}

\author{Boyu Wang$^1$,  Kevin Yager$^2$, Dantong Yu$^2$, and Minh Hoai$^1$ \\
$^1$Stony Brook University, Stony Brook, NY, USA\\
{\tt\small \{boywang, minhhoai\}@cs.stonybrook.edu}\\
$^2$Brookhaven National Laboratory, Upton, NY, USA\\
{\tt\small \{kyager,dtyu\}@bnl.gov}
}
\maketitle
\ifwacvfinal\thispagestyle{empty}\fi

\begin{abstract}

Visual inspection of x-ray scattering images is a powerful technique for probing the physical structure of materials at the molecular scale. In this paper, we explore the use of deep learning to develop methods for automatically analyzing x-ray scattering images. In particular, we apply Convolutional Neural Networks and Convolutional Autoencoders for x-ray scattering image classification. To acquire enough training data for deep learning, we use simulation software to generate synthetic x-ray scattering images. Experiments show that deep learning methods outperform previously published methods by 10\% on synthetic and real datasets.

\end{abstract}

\section{Introduction}


X-ray scattering is used in a wide variety of domains, from determining protein structure to observing realtime structural changes in materials. Broadly speaking, x-ray scattering can probe the physical structure of materials at the molecular and nanoscale. The technique consists of shining a bright, collimated x-ray beam through a material of interest; detailed information about structural order is then inferred from the far-field pattern of scattered rays~\cite{guinier1994x}. The scattering images contain visual features, such as rings, spots, and halos, which encode detailed information about the size, orientation, and packing of atoms, molecules, and nanoscale domains~\cite{yager2014periodic}. Modern x-ray detectors can generate 50,000 to 1,000,000 images/day (1-4 TB/day); thus it is crucial to automate the image processing workflow as much as possible. The lack of 
immediate feedback during x-ray scattering experiments currently limits the scientific productivity of this technique. Manually curated image analysis becomes a bottleneck, due to the enormous diversity of possible image features; we propose instead to develop computer vision algorithms to automate the process of image analysis. 


\begin{figure}
\centering
\includegraphics[width=0.45\textwidth]{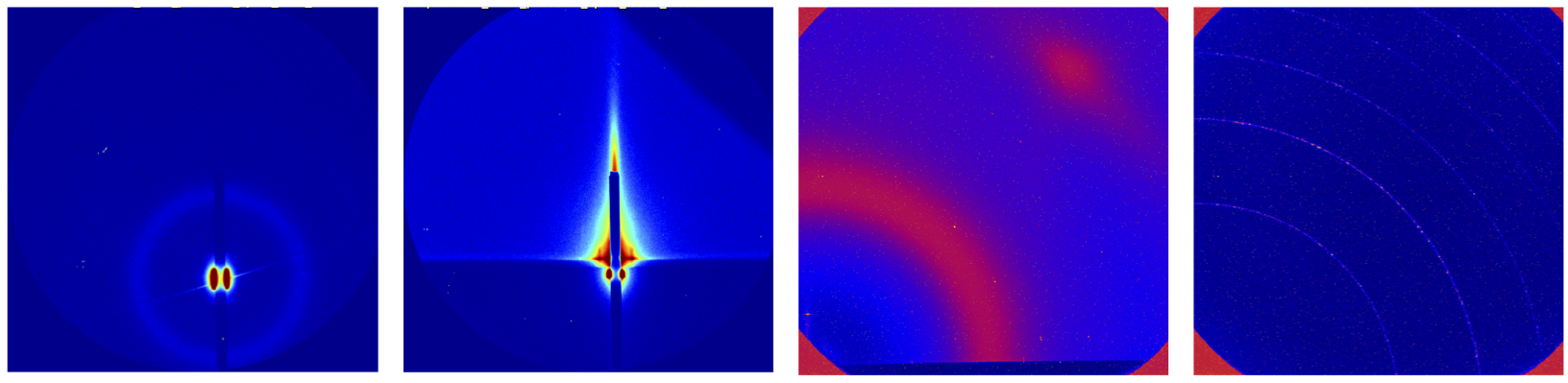}
\vskip 0.1in
\includegraphics[width=0.45\textwidth]{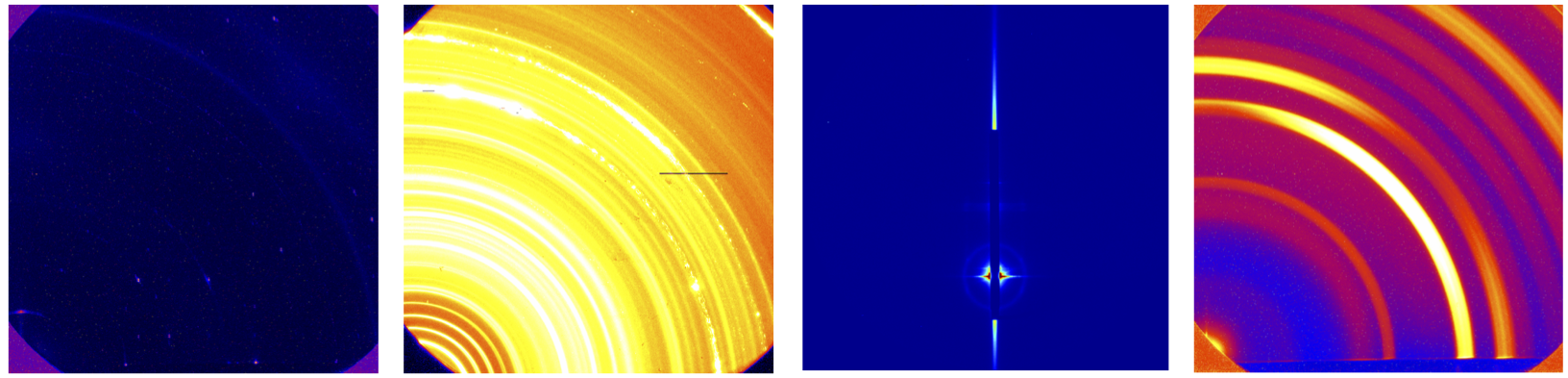}
\caption{Examples of x-ray scattering images. Images are shown using (arbitrary) false-colors; source images are grayscale. We explore the use of deep learning methods to automatically recognize the attributes of x-ray images. The first row of this figure shows some attributes that we want to recognize; from left to right, the attributes are: TSAXS, Linear Beamstop, Halo, and Powder. Automatic attribute recognition is a challenging problem due to the high intraclass variance. Images with the same attribute can look very different; the second row of this figure shows example images with the attribute `ring.' }
\label{fig: ring}
\end{figure}

Our task is to classify x-ray scattering image attributes. These attributes represent a diverse set of characteristics ranging from the type of measurement, e.g. `small-angle x-ray scattering (SAXS)' or `wide-angle x-ray scattering (WAXS)', to instrumental information, e.g. `linear beamstop' or `beam off image',  to appearance-based scattering features, e.g., `halo' or `ring', to chemical composition and physical properties of the materials, e.g., `powder' or `$\mathrm{SiO_2}$'. Figure~\ref{fig: ring} shows example images with the `ring' attribute, which signifies the appearance of a circle or arc of high intensity. As these examples show, images with the same attribute can otherwise be strikingly different. This makes classification of x-ray images extremely challenging.


To the best of our knowledge, there exists little work for automatically analyzing x-ray scattering images. The most closely-related prior work is that of Kiapour et al.~\cite{kiapour2014materials}, which also aimed to recognize the set of image attributes considered in this paper. Their work used hand-designed features such as HOG~\cite{Dalal-Triggs-CVPR05} and SIFT~\cite{Lowe-IJCV04}. Unfortunately, these types of features were designed for natural images instead of x-ray images. As a consequence, the method developed by 
Kiapour et al.~\cite{kiapour2014materials} lacks the performance that would be desired for trustworthy automated analysis of real scientific data. 




In this paper, inspired by the recent success of deep learning methods~\cite{lecun2015deep}, we propose to investigate the use of deep learning for x-ray image classification. In particular, we propose to use Convolutional Neural Networks (CNN)~\cite{LeCun-et-al-NC89,LeCun-et-al-IEEE98, Krizhevsky-et-al-NIPS12} (in particular Residual Networks) and Convolutional Autoencoders~\cite{Hinton-Salakhutdinov-Science06} to extract features that are important for x-ray image classification. Convolutional filters are able to extract local patterns across whole images. With stacking of multiple convolutional layers, CNN is able to extract hierarchical features from images.


However, the size of the previously-available x-ray image dataset is quite small; far too small for robust application of deep learning methods. The dataset collected by~\cite{kiapour2014materials} only contains 2832 images. Manually labeling x-ray images requires a significant amount of 
domain experts' time, which could otherwise be spent on high-level scientific discovery. It becomes infeasible for experts to tag a large dataset containing millions of images. In this paper, we propose a synthetic dataset which is generated based on the known physics underlying x-ray scattering experiments and ad hoc features to emulate the artifacts and defects present in experimental images. By utilizing the synthetic dataset, we can train deep networks for x-ray scattering images. The trained networks can be used to extract feature representations for both synthetic and real x-ray scattering images. Using this type of feature representation, we obtained a  method that outperforms hand-crafted features~\cite{kiapour2014materials}  by 10\% on both synthetic and real datasets.




The rest of the paper is organized as follows: In section~\ref{sec:dataset} we introduce the real and synthetic datasets used in our experiments. In section~\ref{sec: methods}, we describe the deep learning techniques for x-ray scattering image attribute classification. In section~\ref{sec: exps}, we report the performance of our methods. 

\section{Datasets}\label{sec:dataset}
We used two datasets for developing and evaluating deep learning methods for analyzing x-ray scattering images. 

\subsection{X-ray Materials Discovery Dataset (XMD)}

X-ray Materials Discovery Dataset (XMD)~\cite{kiapour2014materials} contains 2832 x-ray scattering images collected from thirteen x-ray scattering measurement runs. The number of images in each experiment varies between 54 and 618. The dataset includes a wide range of samples: nano-particles in solution, lithographic gratings, self-assembling polymer films, organic semiconductors, etc. All images are single-channel with intensities in the range [0, $2^{16}$]. The images have been tagged with 98 attributes by a domain expert. Images are labeled with an average of 11.7 attributes.

Since the images within the same measurement run can be much more similar than across different runs, cross validation should be performed by leaving one measurement run out~\cite{kiapour2014materials}. In other words, we perform 13-fold validation; in each fold, we train the model on all runs except one and use the resulting model to test and predict on the one that is excluded from training. 

Due to the unbalanced distribution of attributes---for example, SAXS and WAXS attributes may be prevalent in many samples however specific material attributes like $SiO_2$ is quite rare---classification accuracy is not a useful evaluation metric. Following~\cite{kiapour2014materials}, we use Average Precision (AP) for evaluation, which is more appropriate for unbalanced data than accuracy.

\subsection{Synthetic Dataset}

\begin{figure*}[t]
\centering
\includegraphics[scale = 0.5]{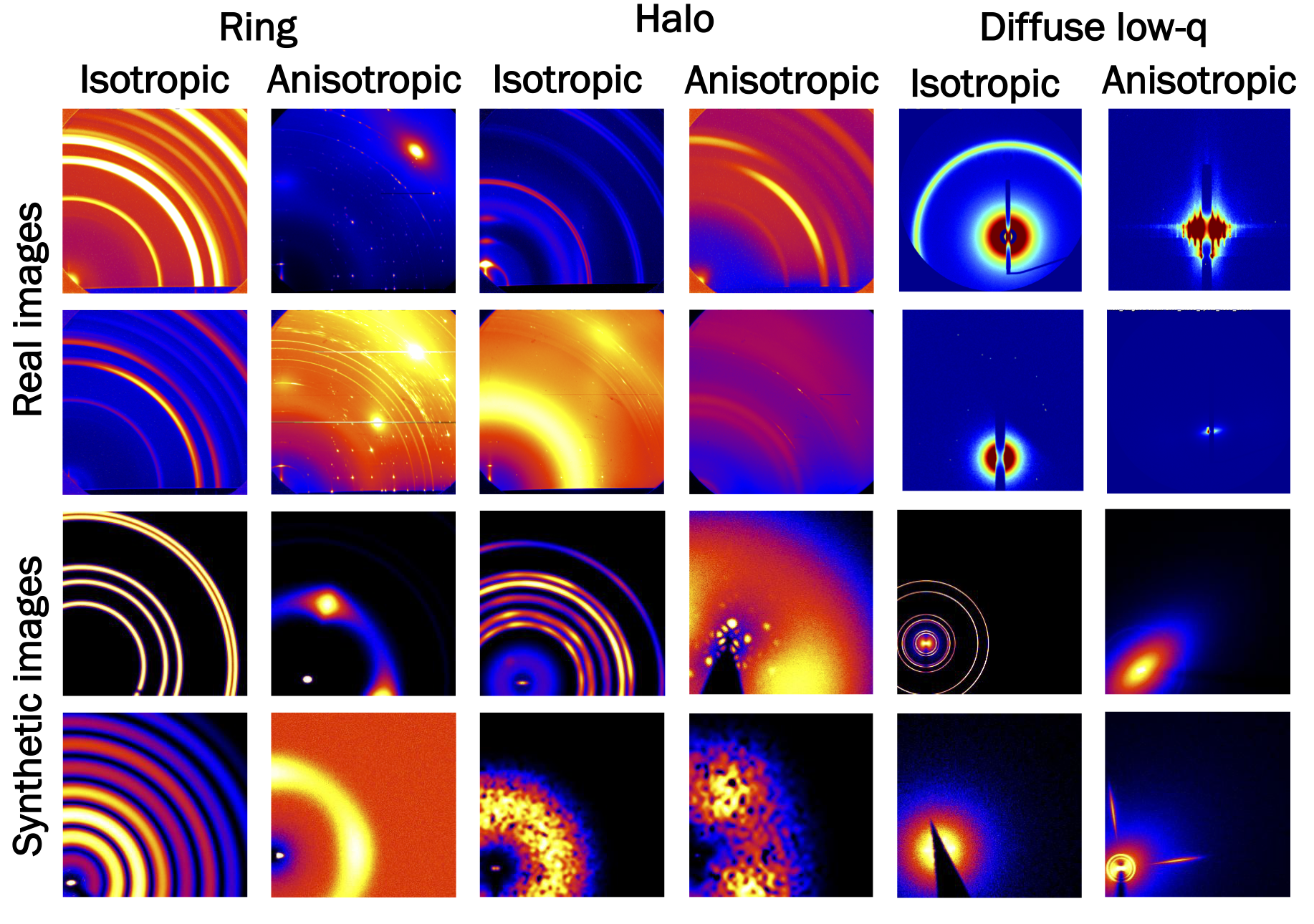}
\caption{Comparison between synthetic images and real experimental images. The first and second rows are real experimental images, while the third and forth rows are synthetic images. Images in the same column have the same attribute. From left to right, the attributes are: Ring: Isotropic, Ring: Anisotropic, Halo: Isotropic, Halo: Anisotropic, Diffuse low q: Isotropic, and Diffuse low q: Anisotropic. Visually, synthetic and real images are indiscernible.
}
\label{fig: synthetic & real}
\end{figure*}  

Deep learning methods generally require large and diverse training sets to yield good performance. Unfortunately, the available human-tagged experimental datasets are very small. To effectively exploit deep learning methods for x-ray scattering image classification, we propose using large datasets with synthetic scattering images. We implemented our own simulation software to generate synthetic datasets by mixing ad-hoc methods and physics-based simulations of scattering data. In the case of ad-hoc images, relevant features such as rings or spots are generated and summed together. In the case of simulations, we combine a variety of well-known methods, including using known analytic models for the expected scattering of certain objects (e.g., spherical nanoparticles) or assemblies (e.g., cubic lattice of objects). The simulation code also includes a module allowing the scattering to be computed for arbitrary arrangements of entities that represent atoms, molecules, or nanoparticles. A particular synthetic image is generated by selecting a random set of simulation modules, and summing together their outputs (based on randomly-selected input variables). The final image is then adjusted to simulate a variety of experimentally-realistic effects, including background noise, shot noise, gaps and shadows arising from experimental geometry. Overall, the simulated images cover a wide space of possibilities, both in terms of the types of observed structures/patterns and the quality of images. 


The simulation code allows the generation of an arbitrary number of training images; where each image carries the appropriate tags (which are known based on the selected simulation modules). In the present experiments, we generate a synthetic dataset which contains 100,000 x-ray scattering images. Figure~\ref{fig: synthetic & real} visualizes the differences between synthetic images and experimental images. We observe that synthetic images and real experimental images are visually similar. This suggests the potential usefulness of synthetic data as training examples. 



\section{Deep-learning Methods}
\label{sec: methods}

In this section, we describe two deep learning techniques to extract features for x-ray scattering images. Rather than using hand-crafted features, which are often designed for natural images, we use  Convolutional Neural Networks to automatically extract features that are important for x-ray images. Once the features are extracted, we train one vs. all support vector machines (SVMs) classifiers~\cite{Vapnik-98} to predict x-ray image attributes. We adopt two methods for feature extraction: one is based on supervised learning, and the other is unsupervised learning. In rest of this section, we will discuss the two methods in details.

\subsection{Residual network}

\begin{figure}
\begin{center}
\includegraphics[scale = 0.23]{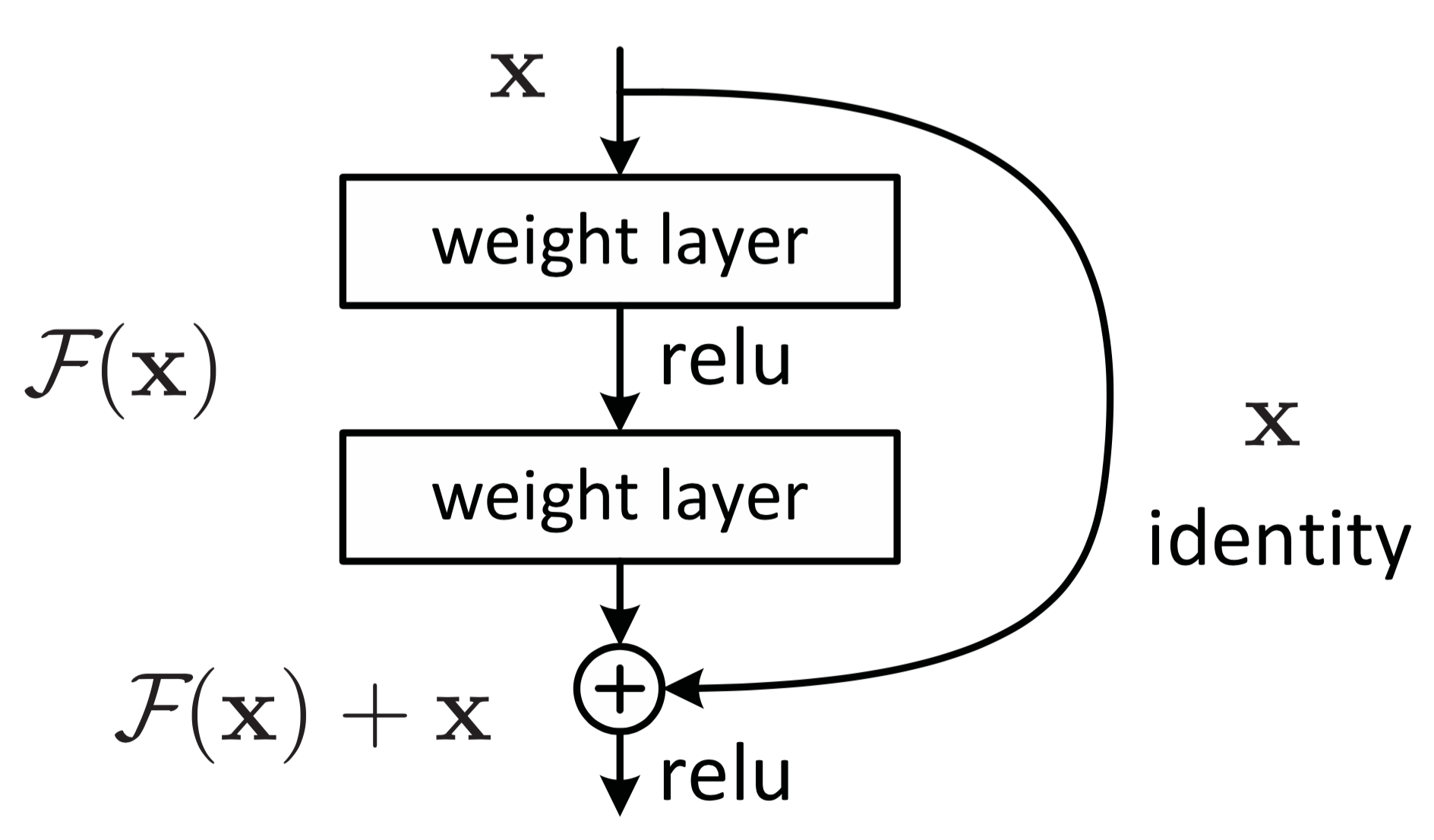}
\end{center}
   \caption{The building block of residual networks~\cite{He2015}. With network bypasses, residual networks explicitly reformulate the layers as learning residual functions with reference to the layer inputs.}
\label{fig: residual network}
\end{figure}

We train a Convolutional Neural Network (CNN) on the synthetic dataset to classify x-ray image attributes. We use the recently proposed 50-layer Residual Network~\cite{He2015} as our architecture. Figure~\ref{fig: residual network} shows the basic learning block in the residual network. Using bypass connections, a residual network explicitly reformulates the layers as learning residual functions with regard to the input layer. By adopting such a framework, deeper networks are easier to optimize than those without bypass connections, gaining accuracy from a considerably deep architecture. Table~\ref{tab: residual network} shows the detailed architecture of our adopted network. We modify the softmax layer to a binary sigmoid layer because image attributes are not mutually exclusive. The dimension of network output is equal to the number of attributes, and each element in the output vector represents the probability of having that attribute. The final loss function is the summation of the losses incurred by each attribute. 

\begin{table}
\begin{center}
\begin{tabular}{l|c|c}
\toprule
layer name & output size  & kernels  \\
\midrule
conv1 & 112$\times$112 & 7$\times$7, 64, stride 2  \\
\hline
\multirow{4}{*}{conv2\_x} & \multirow{4}{*}{56$\times$56} & 3$\times$3 max pool, stride 2 \\
	 \hhline{~~-}
    & & $\left[\kern-\nulldelimiterspace \begin{array}{ *{3}{c} }
     1\times1, 64 \\
     3\times3, 64 \\
     1\times1, 256 \\
    \end{array} \right]$  \rlap{$\times$ 3} \\
    \hline
    
\multirow{3}{*}{conv3\_x} & \multirow{3}{*}{28$\times$28} & 
    $\left[\kern-\nulldelimiterspace \begin{array}{ *{3}{c} }
     1\times1, 128 \\
     3\times3, 128 \\
     1\times1, 512 \\
    \end{array} \right]$  \rlap{$\times$ 4} \\
    \hline  
    
\multirow{3}{*}{conv4\_x} & \multirow{3}{*}{14$\times$14} & 
    $\left[\kern-\nulldelimiterspace \begin{array}{ *{3}{c} }
     1\times1, 256 \\
     3\times3, 256 \\
     1\times1, 1024 \\
    \end{array} \right]$  \rlap{$\times$ 6} \\
    \hline     
        
\multirow{3}{*}{conv5\_x} & \multirow{3}{*}{7$\times$7} & 
    $\left[\kern-\nulldelimiterspace \begin{array}{ *{3}{c} }
     1\times1, 512 \\
     3\times3, 512 \\
     1\times1, 2048 \\
    \end{array} \right]$  \rlap{$\times$ 3} \\
    \hline 
    
pooling & 1$\times$1 & average pooling  \\
\hline
fc & 1$\times$1 & 2048$\times$num of attributes  \\
\bottomrule
\end{tabular}
\end{center}
\vskip -0.1in
\caption{The architecture and parameter settings of the convolutional neural network developed here. It is based on the 50-layer residual network~\cite{He2015}.}
\label{tab: residual network}
\end{table}

We train the residual network on a synthetic dataset with 100,000 x-ray images. During training, we did not train the network to predict the entire set of attributes to avoid the problem of unbalanced data: some attributes appear in the majority of images while some attributes are really rare. We pick a subset of 17 attributes which are not rare and guarantee that every x-ray image has at least one of these 17 attributes. Table~\ref{tab: 17 attributes} lists the 17 selected attributes. Once the network has been trained, we can use it to extract the feature representation for an image patch as follows. First, the image patch is resized to 224$\times$224, and subsequently fed into the network, and secondly, the activation of the network right before the output layer is taken as the feature vector representation. This feature vector has 2048 dimensions, and is conventionally referred to as fc7 feature vector~\cite{Krizhevsky-et-al-NIPS12}. 

\begin{figure}[t]
\includegraphics[width=0.99\linewidth]{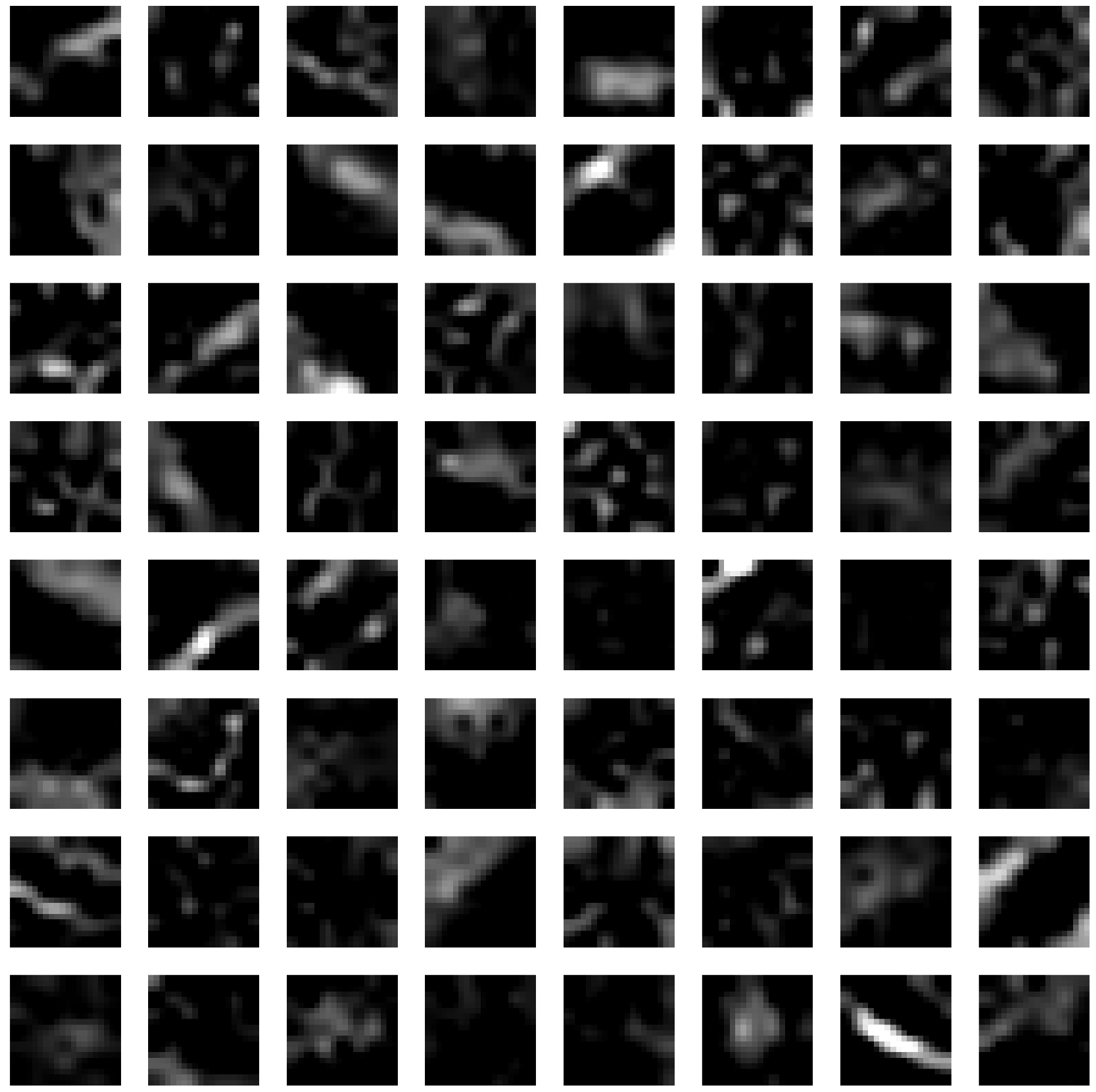}
\vskip -0.1in
\caption{Visualization of the filters of the first layer of the learned residual network.}
\label{fig: filter}
\end{figure}

Figure~\ref{fig: filter} visualizes the first layer's filters learned by our residual network. As can be seen, many filters of the the first layer tend to pick up edge-like signals.

We use the learned network to extract feature vectors for the real experimental XMD dataset. The size of each images in this dataset is 1024$\times$1024, which is larger than the expected input size of the the CNN, which is 224$\times$224. To compute the feature vector representation, we first resize a 1024$\times$1024 image to three different scales: 256$\times$256, 384$\times$384, and 512$\times$512. At each scale, we crop five images of size 224$\times$224 (at the center and four corners), and feed them into the previously-trained network to obtain the corresponding feature vector. We average the corresponding fifteen feature vectors as the final feature representation for the x-ray image.

\begin{table}
\begin{center}
\begin{tabular}{lll}
\toprule
BCC & Beam Off Image & Circ. Beamstop \\
Diffuse high-q & Diffuse low-q & FCC \\
Halo & High background & Higher orders \\
Linear beamstop & Many rings & Polycrystalline \\
Ring & Strong scattering & Structure factor \\
Weak scattering & Wedge beamstop\\
\bottomrule
\end{tabular}
\end{center}
\vskip -0.15in
\caption{The list of 17 attributes for training the residual network. These attributes were selected to be representative of the diversity of labels associated with x-ray scattering images.}
\label{tab: 17 attributes}
\end{table}

\subsection{Convolutional Autoencoder network}

\begin{figure}[t]
\includegraphics[width=0.99\linewidth]{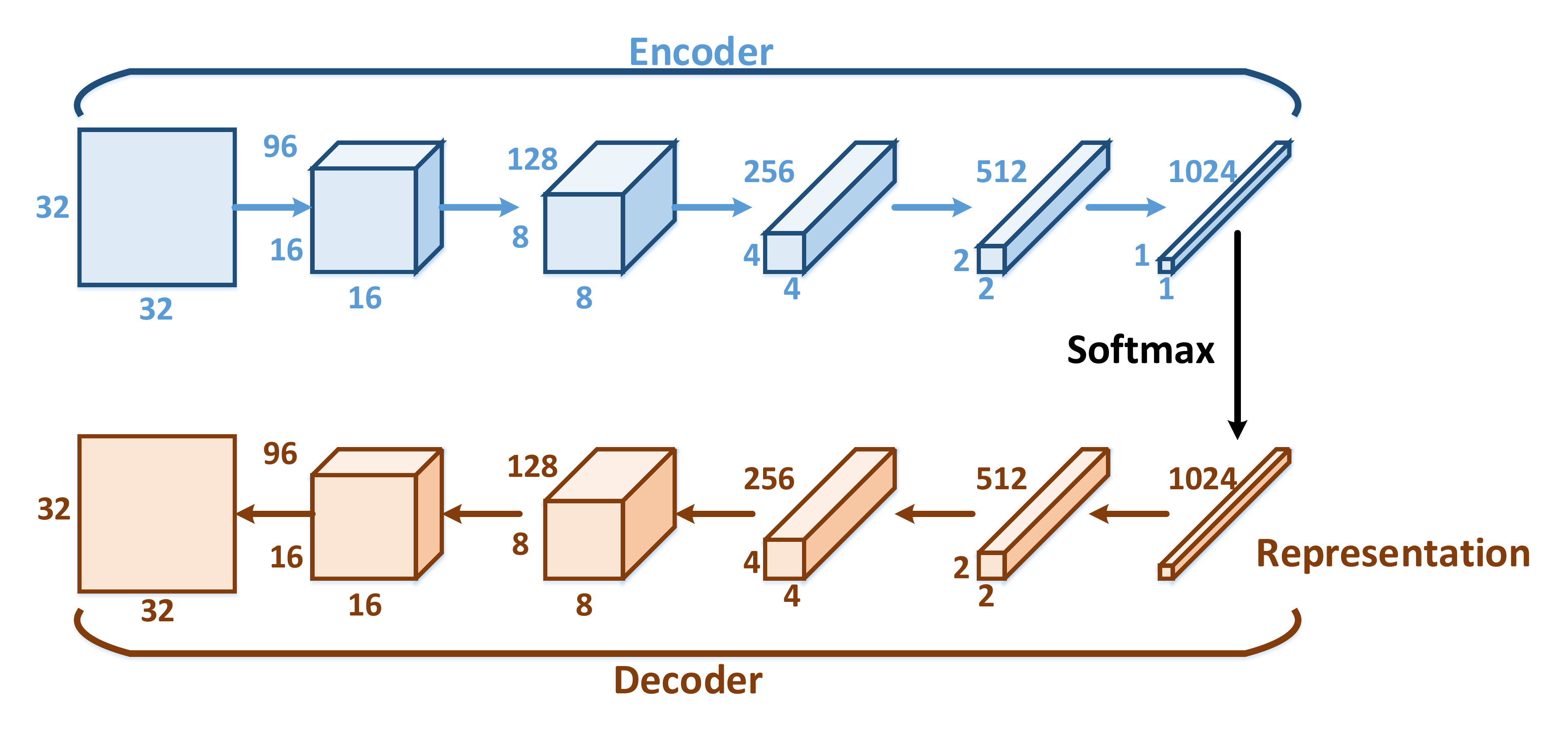}
\vskip -0.1in
\caption{Architecture of the convolutional autoencoder. The difference between this architecture and the traditional autoencoder is that we add a softmax layer after the output of encoder and we use this vector, of which the size is 1024, as the feature vector for an image patch.}
\label{fig: autoencoder}
\end{figure}

The second method to extract feature vectors is the convolutional autoencoder. Instead of performing autoencoding on full-size images, we perform autoencoding on image patches because: i) the size of original images is too big to train an autoencoder; and ii) downsampling original image to a smaller resolution such as $256\times256$ or  $128\times128$ may lose some important details, such as sharp, localized peaks. Since we train the autoencoder at the patch level, even though we have a limited number of real experimental images (2832 images in XMD dataset), we still obtain many image sample patches from each image.  Therefore, our convolutional autoencoder network is trained on real experimental image patches. We resize 1024$\times$1024 images to multiple scales and randomly extract 1000 32$\times$32 patches per image as the training set for autoencoder. Figure~\ref{fig: autoencoder} shows the architecture of our autoencoder for image patches. The difference between our autoencoder and traditional autoencoder is that we applied a softmax layer after the output of encoder and used the output vector with the size of 1024 as the feature vector for input image patch. The objective here is to use the autoencoder to cluster image patches. The dimension of the encoder output represents the number of clusters, and each element belongs to a data cluster. After applying the softmax function, the vector represents the possibilities that the input image falls into different data clusters. 

\begin{figure}[t]
\centering
\includegraphics[width=0.98\linewidth]{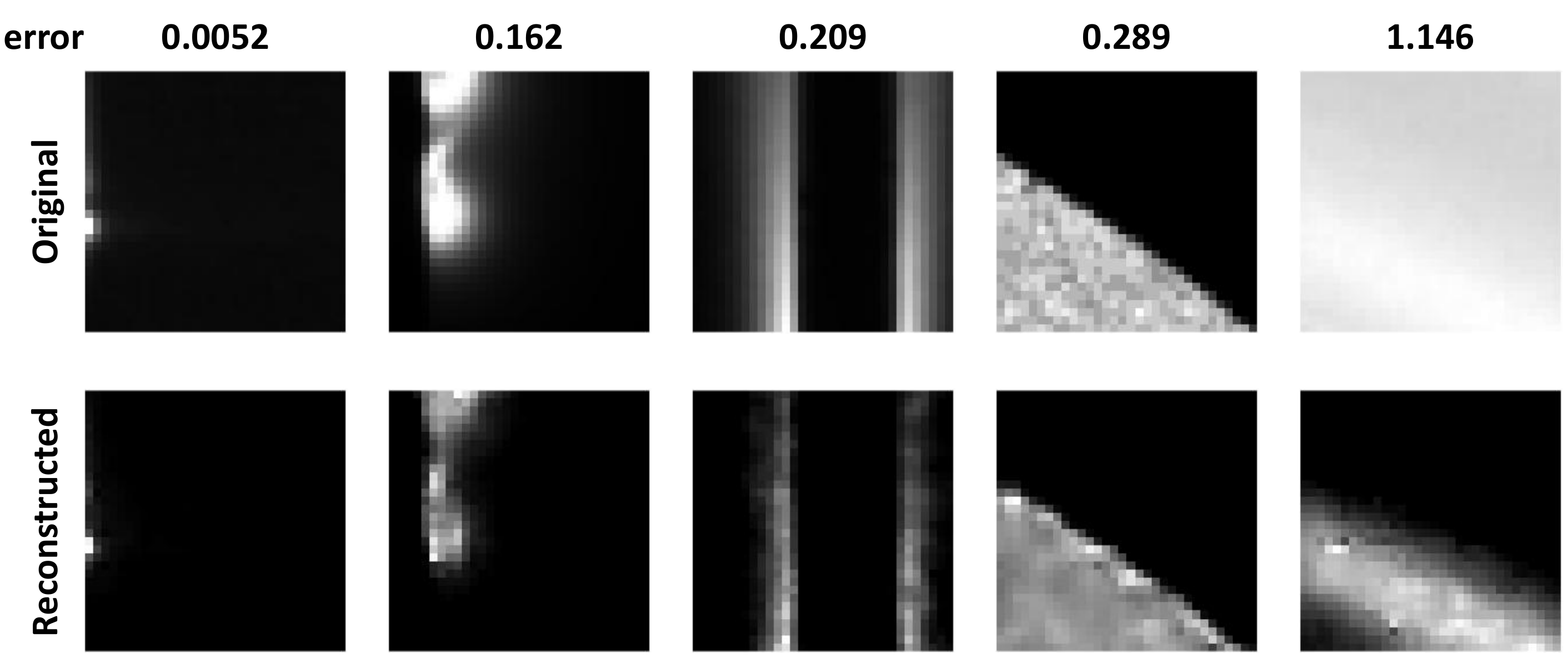}
\caption{Visualization of original image patches (top row) and reconstructed image patches (bottom row) using convolutional autoencoder. From left to right, the reconstruction errors are 0.0052, 0.162, 0.209, 0.289 and 1.146. Here the reconstructed images retain the important visual structures of the original images even when the reconstruction error is high (at the right most column).}
\label{fig: autoencoder reconstruction}
\end{figure}

In our experiment, the learned autoencoder has the minimum reconstruction error of 0.0044, the maximum of 1.2510, and the average of 0.6817. Figure~\ref{fig: autoencoder reconstruction} shows the original in the top row and reconstructed image patches at the bottom row. For left to right, the reconstruction error (the difference between the top image and the bottom image) increases. The reconstruction retains the important visual structures of the original images.

\begin{figure*}[t]
\centering
\includegraphics[scale = 0.65]{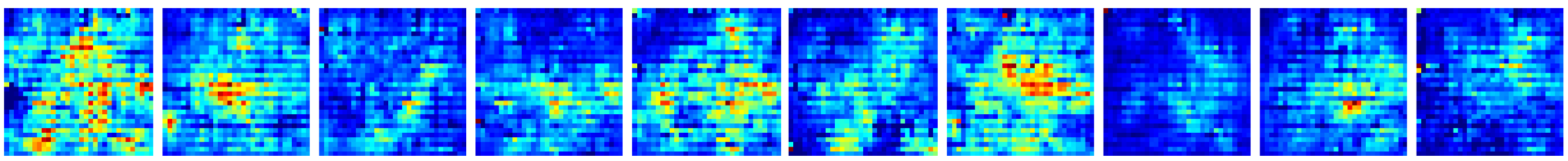}
\vskip -0.1in
\includegraphics[scale = 0.65]{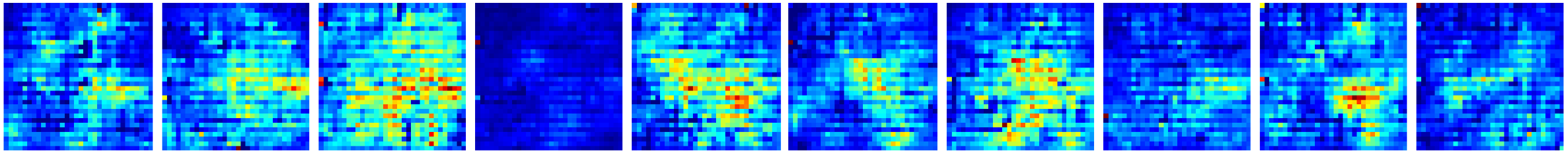}
\caption{Visualization of clusters. These data clusters capture edges of different angle, and blobs of different location.}
\label{fig: autoencoder cluster}
\end{figure*}

Representation vectors have 1024 dimensions. By setting the representation vector to be one-hot vector (all values are 0 expect one location is 1), we get the data cluster represented in that location. By changing the location, we can get 1024 different data clusters. Figure~\ref{fig: autoencoder cluster} visualizes 20 data clusters. These data clusters capture the edges with different angles and the blobs at different locations.

After training the autoencoder, we perform a spatial pyramid matching (SPM)~\cite{Lazebnik-et-al-CVPR06} with three levels to extract the feature vector descriptor for an image. The SPM partitions an image into sub-regions of increasingly fine granularity and computes the histograms of local features found within each sub-region. This approach allows the classifier to better understand the spatial relationship between different areas of an image. We use sum pooling because we compute histograms and intent to obtain the frequency of each cluster. 

\section{Experimental results}
\label{sec: exps}
In this section, we report the performance of the deep-learning features for x-ray image classification. We report the average precision values on both synthetic and real datasets, and compare them with previously published results. 


\subsection{Performance on synthetic dataset}
We first evaluate the performance of the residual network on the holdout (test subset) of the synthetic dataset (note that we split the synthetic dataset into two disjoint train/test subsets). For the test data, the residual network achieves the mean average precision of 77.1\% for the 17 attributes given in Table~\ref{tab: syn performance} (note that we trained the residual network on  these 17 attributes only). The result is given in Table~\ref{tab: syn performance}. We compare this result with a shallow-learning method that is based on the Bag-of-Word approach~\cite{Leung-Malik-IJCV01,Sivic-Zisserman-ICCV03}. This approach consists of: i) using $k$-means to learn a visual codebook; ii) applying spatial pyramid pooling to generate a feature vector representation for each image; iii) training  a binary SVM for each of 17 attributes in consideration. This approach only achieves a mean Average Precision of 67.1\%, which is 10\% lower than our method using the feature descriptors from deep-learning. This indicates the benefits of using deep-learning for x-ray image classification. 



\begin{table}
\begin{center}
\begin{tabular}{lll}
\toprule
Method & mAP  \\
\midrule
$k$-means + Bag-of-Words  + Spatial Pyramid &  67.1 \\
Residual network (deep learning) &  77.1 \\
\bottomrule
\end{tabular}
\end{center}
\vskip -0.1in
\caption{Mean Average Precision (mAP) on synthetic dataset for 17 attributes}
\label{tab: syn performance}
\end{table}

\subsection{Performance on XMD dataset}
For XMD dataset, we use both the residual network and the convolutional autoencoder to extract features for each image, then train a multi-class SVM~\cite{REF08a} with a one vs. all approach on all 98 attributes. Following~\cite{kiapour2014materials}, we report the performance on these 98 attributes in Table~\ref{tab: map 98 tags}. The residual network features achieve 59.5\% mean AP, which outperforms hand-crafted features by 8\%. The convolutional autoencoder features achieve a mean AP of 57\%. Combining the features generated both the residual network and convolutional autoencoder,  we obtain the mean AP of 61.1\%, outperforming hand-crafted features by 10\%. In~\cite{kiapour2014materials} they also report the performance using a hierarchical classifier in which the first classifier is used to separate images into two categories: small-angle (SAXS) and  wide-angle (WAXS), and subsequently the second classifier is used to classify the fine-grained attributes in each category. The hierarchical approach boosts the performance from 51.5\% to 55.5\%. In our method, we have not used the hierarchical approach, and we hypothesize that the hierarchical approach will also improve the performance of our method. One of our future works is to build a more fine grained hierarchical structure based on attributes correlations to boost the performance. 
  
\begin{figure}[t]
\centering
\begin{subfigure}[b]{0.23\textwidth}
   	\includegraphics[width=\textwidth]{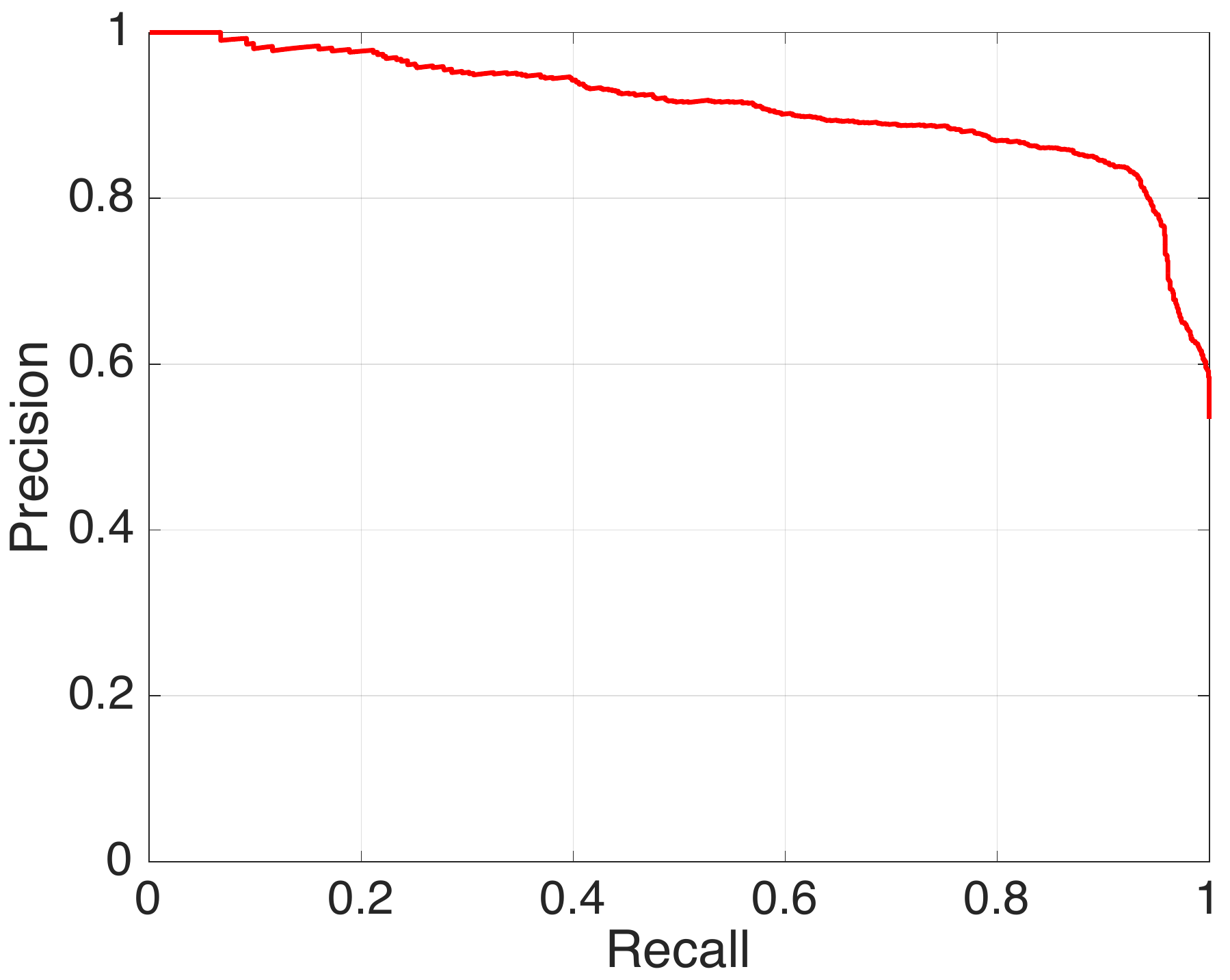}
 	\caption{Ring}
 	\end{subfigure}
   \begin{subfigure}[b]{0.23\textwidth}
   	\includegraphics[width=\textwidth]{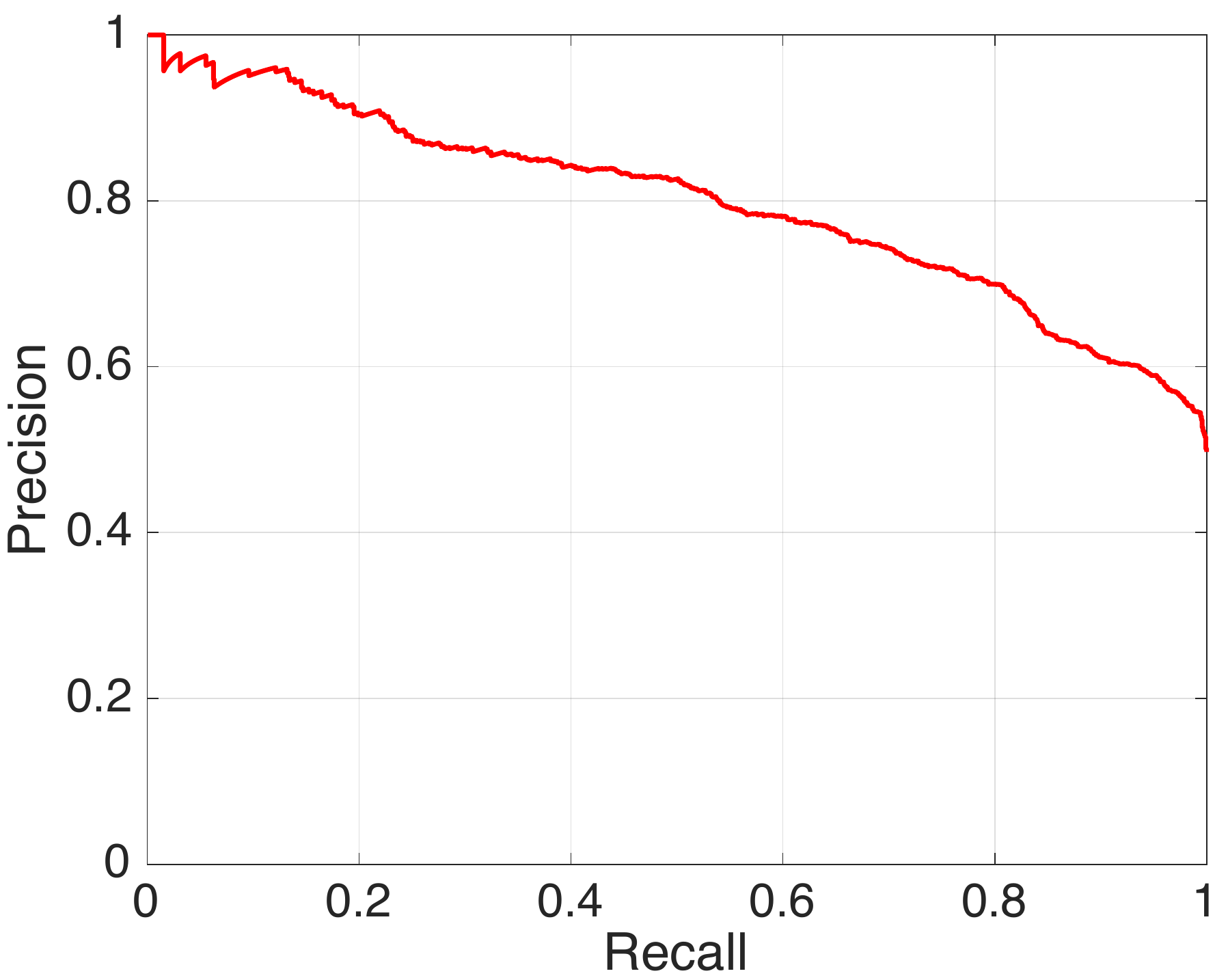}
    \caption{Peaks}
     \end{subfigure}
     \vskip -0.1in
\caption{Precision-recall curves for recognizing attributes `Ring' and `Peaks'.  } 
\label{fig: prc}
\end{figure}

Analyzing the XMD data, we discovered that 18 out of 98 attributes only appear in a single experiment run (of the 13 experiment runs). Therefore, for the leave-one-experiment-out cross validation procedure, these attributes will only appear in either the training set or test set. When an attribute does not appear in the test set, the average precision for this attribute is one. When the attribute does not appear in the train set, the average precision is equal to the average precision of a random classifier. In both cases, the results do not depend on the type of the feature representation. Herein, we suggest to remove these attributes from the experimental analysis to obtain a more indicative mean average precision value. The mean average precision on the 80 remaining attributes is shown in Table~\ref{tab: map 80 tags}. Figure~\ref{fig: prc} shows precision-recall curves of the method that uses residual network features for recognizing `rings' and `peaks'.


\begin{table}
\begin{center}
\begin{tabular}{lll}
\toprule
Feature & mAP  \\
\midrule
lbpphog~\cite{kiapour2014materials} & 51.5 \\
Residual network & 59.5 \\
Patch autoencoder & 57.9\\
Residual network + Patch autoencoder & 61.1  \\
\bottomrule
\end{tabular}
\end{center}
\vskip -0.15in
\caption{Mean Average Precision (mAP) for all 98 attribute classifiers on the XMD dataset. Deep-learning features (Residual network and Path autoencoder) outperform hand-designed features by a large margin. The best result is achieved when two types of deep-learning features are combined.}
\label{tab: map 98 tags}
\end{table}

\begin{table}
\begin{center}
\begin{tabular}{lll}
\toprule
Feature & mAP  \\
\midrule
Residual network & 59.03 \\
Patch autoencoder & 56.00\\
Residual network + Patch autoencoder & 59.95  \\
\bottomrule
\end{tabular}
\end{center}
\vskip -0.15in
\caption{Mean Average Precision (mAP) for 80 attributes. Each attribute in this experiment appears in at least two experiment runs of the XMD dataset.}
\label{tab: map 80 tags}
\end{table}
 

\begin{figure}[t]
\centering
\includegraphics[width=0.95\linewidth]{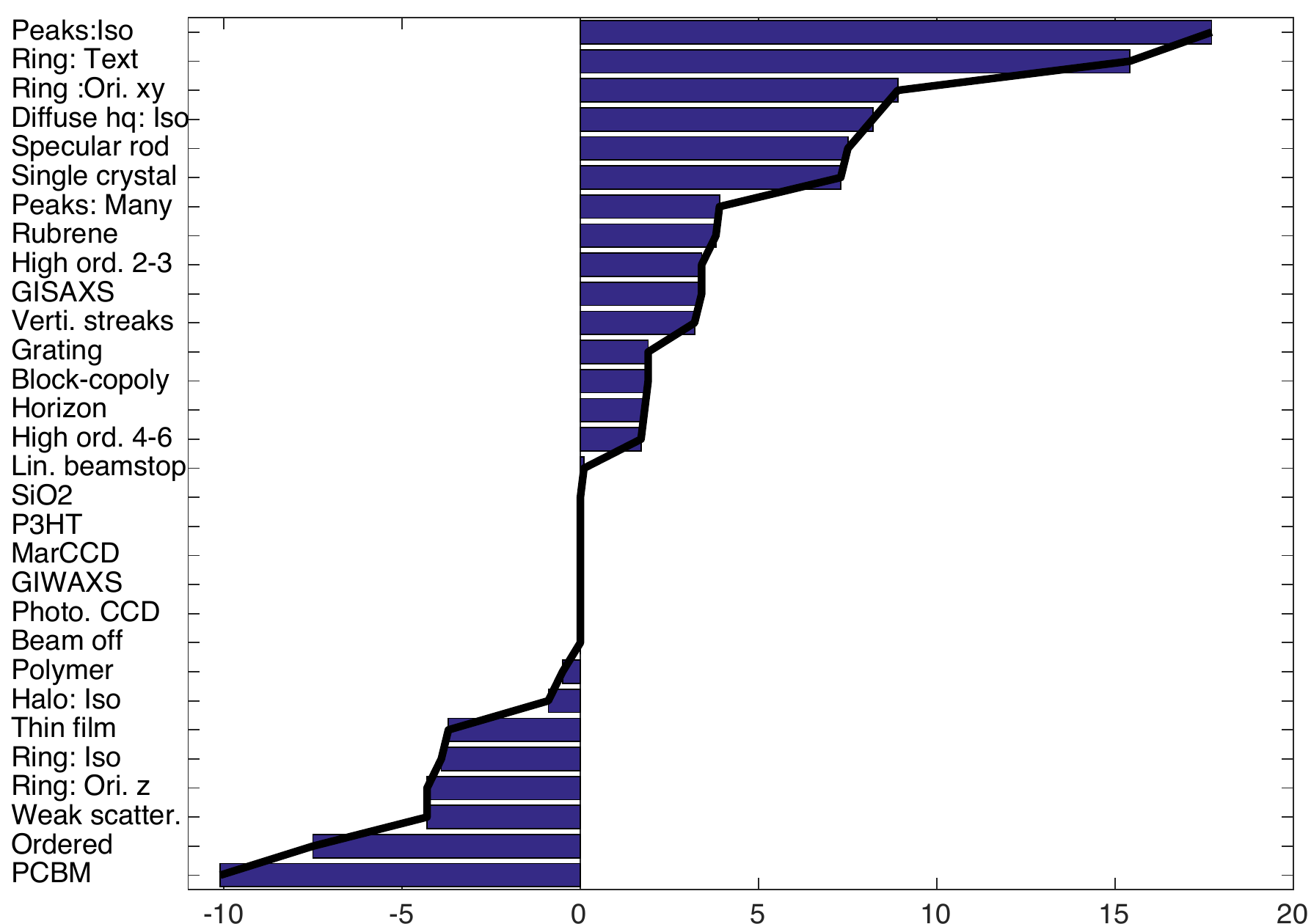}
\caption{{\bf Average precision (AP) gap.} This shows the difference between the average precision of deep learning features and hand-crafted features. The AP of deep learning features are higher than or equal to the AP of hand-crafted features for the majority of attributes.} 
\label{fig: gap}
\end{figure}

We also compared the performance of using  hand-crafted features and  deep learning features for 30 attributes of which the APs are explicitly reported in~\cite{kiapour2014materials}.  We consider the AP gap, which is the difference between the AP obtained by using deep learning features and the AP obtained by the hand-crafted Ibpphog~\cite{kiapour2014materials}. 
Figure~\ref{fig: gap} shows AP gap for all 30 attributes. From the figure, we observe that the deep learning features outperform the hand-crafted features by a large margin for detecting multiple attributes, including Specular rod, Peaks: Isotropic, Ring: Textured, High order 2-3, Ring: oriented xy, Vertical streaks, Single crystal, Block-cropoly, Peaks: Many, Grating, Diffuse high-q: Isotropical, High order 4-6, and Rubrene. However, for the attributes such as Ordered, Ring: oriented z, Ring: Isotropic, PCBM, and Weak Scattering, the deep learning features do not perform as well as the hand-crafted features. We note that both approaches (deep learning and hand-crafted) perform  well for image attributes where the possible options are disjoint and highly distinct. For instance, detecting the type of detector used in the experiment (MarCCD vs. Photonics CCD) is an `easy' task for both approaches. The hand-crafted approach appears to slightly out-perform deep learning on a small number of attributes, especially those that denote a rather vague interpretation of the overall image (e.g., ordered). Conversely, deep learning achieves a much-improved performance for attributes denoting distinct localized features such as Specular rod and Peaks: Isotropic. Importantly, deep learning can evidently identify the attributes associated with combining a number of disparate features throughout image such as single crystal; that is, we confirm that the hierarchical, multi-level internal representations of deep learning are well aligned with the complex, multi-feature labels that are frequently used by domain experts to describe x-ray scattering images.

\begin{figure}
\centering
\includegraphics[width=0.98\linewidth]{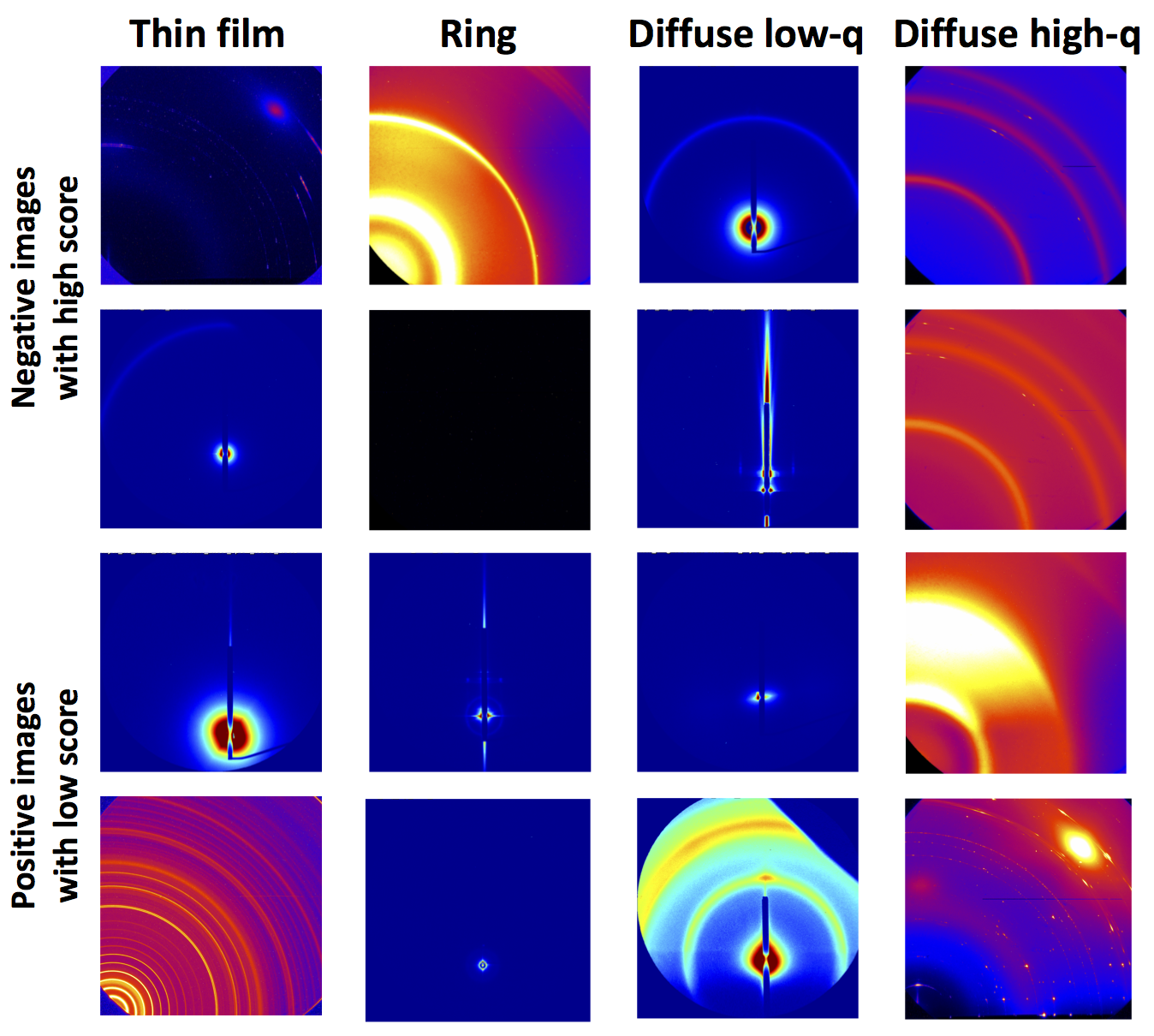}
\vskip -0.1in
\caption{Example images of mismatch between the prediction of our methods, and the human-assigned tag. The first two rows show negative images with high score (i.e., a human annotator tags this image as negative for this attribute, but our prediction method predicts high scores for this attributes). The third and fourth rows show positive images with low score (i.e., a human annotator tags this image as positive for this attribute, but our prediction method predicts low scores for this attributes). Attributes from left to right are: Thin film, Ring, Diffuse low-q, Diffuse high-q. } 
\label{fig: false}
\end{figure}

Figure~\ref{fig: false} shows some images where our method using deep-learning features do not agree with human annotation. The first two rows show some false positives (which means a human annotator tags this image as negative for this attribute, but our method predict it with a high scores for this attribute). The third and fourth rows show some false negative results (which means a human annotator tags this image as positive for this attribute, but our method labels it with a low score for this attribute). After consulting with domain experts, images in the first row reflect some human annotation error in tagging original images, i.e., our method successfully identified human annotation errors. Some images are ambiguous even for a human to detect, e.g., diffuse high-q scattering is weak and spatially distributed. Even human experts may disagree with respect to the marginal (weakly-scattering) examples of this feature. Nevertheless, our approach failed to classify some images because the target attribute has unusual appearance or is highly localized: for example, the positive image with the `Ring' attribute the the third row and second column. It is rare to see a full ring appearing in this type of experiment (GISAXS). The lack of training examples potentially explains why this atypical scattering pattern was mis-classified.   The image on the third row and first column is another mis-classified example that is rare and contains a positive `Thin film' attribute. This is indeed a measurement of a thin film. However, the common visual features, such as a distinct horizontal stripe, are not present possibly because of the misalignment during the experiment.  Only the subtle hints exist to help a trained human expert correctly classify it as a thin film measurement. There is a broad class of images that even a human expert finds extremely challenging to classify, owing to the subtlety of the features, the ambiguity of the tag, or the violation of standard experimental assumptions (e.g., an error occurs during experiments).  Consequently these types of images are challenging to any deep learning based classification approach. By augmenting training sets with some examples of borderline cases, we can improve the classification accuracy for these atypical images. Thus, one potential future work is to augment our simulation code to generate synthetic images exhibiting a variety of atypical or marginal patterns.

\section{Conclusions}
\label{sec: analysis}
We have explored the use of deep learning methods for automatic recognition of x-ray scattering images' attributes. To overcome the size limitation of available annotated x-ray image datasets, we used simulation software to generate synthetic x-ray images for training.  Our features are based on fully connected layer output of a residual network and the representation layer of a convolution autoencoder. Evaluations on both synthetic and real datasets show that deep-learning features outperform hand-crafted features by a large margin of 10\%, using mean average precision as the evaluation metric.


{\small
\bibliographystyle{ieee}
\bibliography{egbib,shortstrings,pubs}

\begin{thebibliography}{10}\itemsep=-1pt

\bibitem{Dalal-Triggs-CVPR05}
N.~Dalal and B.~Triggs.
\newblock Histograms of oriented gradients for human detection.
\newblock In {\em Proc. CVPR}, 2005.

\bibitem{REF08a}
R.-E. Fan, K.-W. Chang, C.-J. Hsieh, X.-R. Wang, and C.-J. Lin.
\newblock {LIBLINEAR}: A library for large linear classification.
\newblock {\em Journal of Machine Learning Research}, 9:1871--1874, 2008.

\bibitem{guinier1994x}
A.~Guinier.
\newblock {\em X-ray diffraction in crystals, imperfect crystals, and amorphous
  bodies}.
\newblock Courier Corporation, 1994.

\bibitem{He2015}
K.~He, X.~Zhang, S.~Ren, and J.~Sun.
\newblock Deep residual learning for image recognition.
\newblock {\em arXiv preprint arXiv:1512.03385}, 2015.

\bibitem{Hinton-Salakhutdinov-Science06}
G.~Hinton and R.~Salakhutdinov.
\newblock Reducing the dimensionality of data with neural networks.
\newblock {\em Science}, 313(5786):504--507, 2006.

\bibitem{kiapour2014materials}
M.~H. Kiapour, K.~Yager, A.~C. Berg, and T.~L. Berg.
\newblock Materials discovery: Fine-grained classification of x-ray scattering
  images.
\newblock In {\em IEEE Winter Conference on Applications of Computer Vision},
  pages 933--940. IEEE, 2014.

\bibitem{Krizhevsky-et-al-NIPS12}
A.~Krizhevsky, I.~Sutskever, and G.~Hinton.
\newblock {ImageNet} classification with deep convolutional neural networks.
\newblock In {\em NIPS}, 2012.

\bibitem{Lazebnik-et-al-CVPR06}
S.~Lazebnik, C.~Schmid, and J.~Ponce.
\newblock Beyond bags of features: Spatial pyramid matching for recognizing
  natural scene categories.
\newblock In {\em Proc. CVPR}, 2006.

\bibitem{lecun2015deep}
Y.~LeCun, Y.~Bengio, and G.~Hinton.
\newblock Deep learning.
\newblock {\em Nature}, 521(7553):436--444, 2015.

\bibitem{LeCun-et-al-NC89}
Y.~LeCun, B.~Boser, J.~S. Denker, and D.~Henderson.
\newblock Backpropagation applied to handwritten zip code recognition.
\newblock {\em Neural Computation}, 1(4):541--551, 1989.

\bibitem{LeCun-et-al-IEEE98}
Y.~LeCun, L.~Bottou, Y.~Bengio, and P.~Haffner.
\newblock Gradient-based learning applied to document recognition.
\newblock {\em Proceedings of the IEEE}, 86(11):2278--2324, 1998.

\bibitem{Leung-Malik-IJCV01}
T.~Leung and J.~Malik.
\newblock Representing and recognizing the visual appearance of materials using
  three-dimensional textons.
\newblock {\em IJCV}, 43(1):29--44, 2001.

\bibitem{Lowe-IJCV04}
D.~Lowe.
\newblock Distinctive image features from scale-invariant keypoints.
\newblock {\em IJCV}, 60(2):91--110, 2004.

\bibitem{Sivic-Zisserman-ICCV03}
J.~Sivic and A.~Zisserman.
\newblock Video {Google}: A text retrieval approach to object matching in
  videos.
\newblock In {\em Proc. ICCV}, 2003.

\bibitem{Vapnik-98}
V.~Vapnik.
\newblock {\em Statistical Learning Theory}.
\newblock Wiley, New York, NY, 1998.

\bibitem{yager2014periodic}
K.~G. Yager, Y.~Zhang, F.~Lu, and O.~Gang.
\newblock Periodic lattices of arbitrary nano-objects: modeling and
  applications for self-assembled systems.
\newblock {\em Journal of Applied Crystallography}, 47(1):118--129, 2014.

\end{thebibliography}
}

\end{document}